\documentclass[10pt,twocolumn,letterpaper]{article}

\usepackage{cvpr}              

\usepackage[dvipsnames]{xcolor}

\definecolor{cvprblue}{rgb}{0.21,0.49,0.74}

\usepackage[accsupp]{axessibility} 

\usepackage[pagebackref,breaklinks,colorlinks,citecolor=cvprblue]{hyperref}
\usepackage{amsmath}
\usepackage{graphicx}

\usepackage[most]{tcolorbox}

\definecolor{mplblue}{rgb}{0.12156862745098039, 0.4666666666666667, 0.7058823529411765}
\definecolor{mplorange}{rgb}{1.0, 0.4980392156862745, 0.054901960784313725}
\definecolor{mplgreen}{rgb}{0.17254901960784313, 0.6274509803921569, 0.17254901960784313}
\definecolor{mplred}{rgb}{0.8392156862745098, 0.15294117647058825, 0.1568627450980392}
\definecolor{mplpurple}{rgb}{0.5803921568627451, 0.403921568627451, 0.7411764705882353}
\definecolor{mplbrown}{rgb}{0.5490196078431373, 0.33725490196078434, 0.29411764705882354}
\definecolor{mplpink}{rgb}{0.8901960784313725, 0.4666666666666667, 0.7607843137254902}
\definecolor{mplgray}{rgb}{0.4980392156862745, 0.4980392156862745, 0.4980392156862745}
\definecolor{mplyellow}{rgb}{0.7372549019607844, 0.7411764705882353, 0.13333333333333333}
\definecolor{mplcyan}{rgb}{0.09019607843137255, 0.7450980392156863, 0.8117647058823529}

\usepackage{multirow}
\usepackage{hhline}
\usepackage{pifont}  
\usepackage{wrapfig}

\def\paperID{10388} 
\def\confName{CVPR}
\def\confYear{2024}

\title{Visual Fact Checker: Enabling High-Fidelity Detailed Caption Generation}

\author{
Yunhao Ge~~~~Xiaohui Zeng~~~~Jacob Samuel Huffman~~~~Tsung-Yi Lin~~~~Ming-Yu Liu~~~~Yin Cui\\\\
NVIDIA\\
\tt\normalsize\url{https://research.nvidia.com/labs/dir/vfc/}
}

\begin{document}

\twocolumn[{%
\renewcommand\twocolumn[1][]{#1}%
\maketitle
\begin{center}
    \centering
    \captionsetup{type=figure}
    \includegraphics[width=1.0\textwidth]{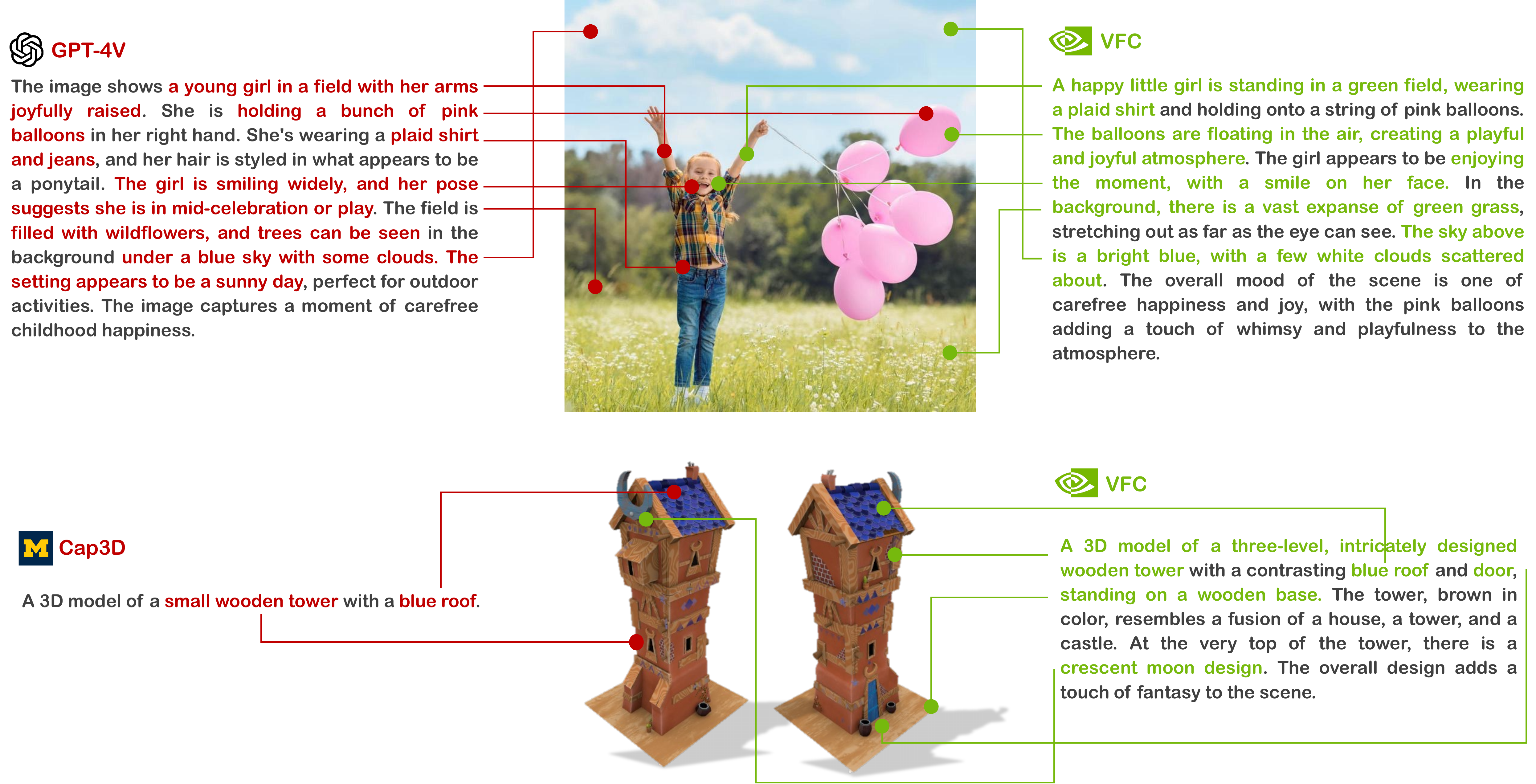}
    \captionof{figure}{Comparison of VisualFactChecker (VFC) with GPT-4V and Cap3D. VFC can generate high-fidelity detailed captions that closely match GPT-4V's quality for 2D images and offer significantly more details for 3D objects than Cap3D. VFC used a pre-trained Llama-2 as the LLM when generating the caption for the above 2D image.}
    \label{fig:teaser}
\end{center}%
}]
\begin{abstract}
Existing automatic captioning methods for visual content face challenges such as lack of detail, content hallucination, and poor instruction following. In this work, we propose VisualFactChecker (VFC), a flexible training-free pipeline that generates high-fidelity and detailed captions for both 2D images and 3D objects. VFC consists of three steps: 1) proposal, where image-to-text captioning models propose multiple initial captions; 2) verification, where a large language model (LLM) utilizes tools such as object detection and VQA models to fact-check proposed captions; 3) captioning, where an LLM generates the final caption by summarizing caption proposals and the fact check verification results. In this step, VFC can flexibly generate captions in various styles following complex instructions. We conduct comprehensive captioning evaluations using four metrics: 1) CLIP-Score for \emph{image-text} similarity; 2) CLIP-Image-Score for measuring the \emph{image-image} similarity between the original and the reconstructed image generated by a text-to-image model using the caption. 3) human study on Amazon Mechanical Turk; 4) GPT-4V for \emph{fine-grained} evaluation. Evaluation results show that VFC outperforms state-of-the-art open-sourced captioning methods for 2D images on the COCO dataset and 3D assets on the Objaverse dataset. Our study demonstrates that by combining open-source models into a pipeline, we can attain captioning capability comparable to proprietary models such as GPT-4V, despite being over 10$\times$ smaller in model size.
\end{abstract}    
\vspace{-6mm}
\section{Introduction}
\label{sec:intro}

Image captioning is a pivotal challenge in computer vision and natural language processing. Its central goal is to encapsulate visual data within a textual description, which requires a nuanced understanding of both modalities. 
The recent advent of multimodal large language models (MM-LLMs), such as GPT-4V~\citep{openai2023gpt4}, and text-to-image generation models, such as DALLE-3~\citep{dalle3}, has marked significant progress in this field. These proprietary models could leverage expansive human-labeled data and enormous computing resources to learn to generate detailed and contextually appropriate image descriptions. 
On the other hand, existing open-sourced captioning methods in the community still face significant challenges. 
Methods such as BLIP-2~\citep{li2023blip} and OFA~\citep{wang2022ofa} often yield overly succinct captions that neglect essential visual information. Conversely, systems like Mini-GPT4~\citep{zhu2023minigpt}, InstructBLIP~\citep{instructblip}, and LLaVA~\citep{liu2023visual, liu2023improved} can suffer from hallucination, producing long descriptions that do not align with the actual content of the images.

\begin{figure*}[t]
\centering
\includegraphics[width=\textwidth]{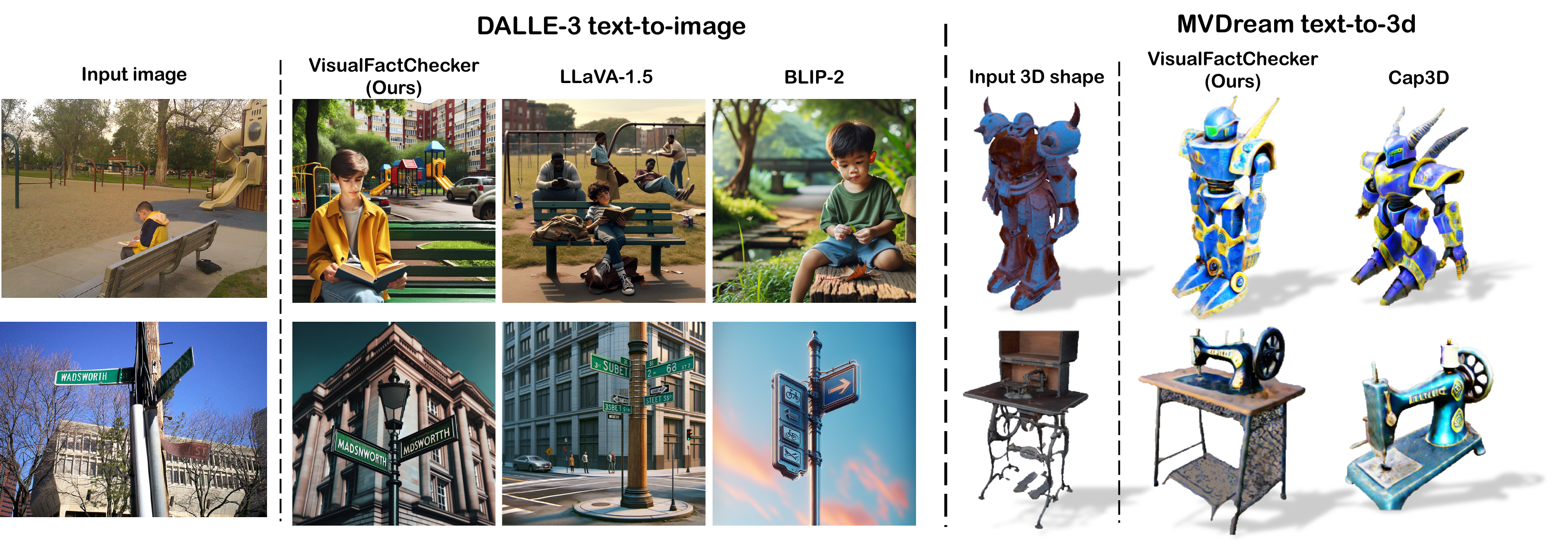}
\caption{We use DALLE-3~\citep{dalle3} as a text-to-image model to reconstruct 2D images using generated captions from different captioning methods (BLIP-2, LLaVA-1.5 and ours). Similarly, we use MVDream~\citep{shi2023mvdream} as a text-to-3D model to reconstruct 3D objects using different 3D captions (generated by Cap3D~\citep{luo2023cap3d} and ours). From the results, we can see that the reconstructed images or 3D objects using BLIP-2 or Cap3D captions are less similar than the input ones, suggesting their captions may not contain sufficient information or incorrectly describe the visual contents; the reconstructed images using LLaVA-1.5 captions contain objects or scenes that are not present in the original images (top: people in the background, bottom: pedestrians and cars on the street), suggesting there might be hallucinations in LLaVA-1.5 captions. Images or 3D objects reconstructed using our captions are more similar to the inputs.}
\label{fig:image-recon}
\end{figure*}

In light of this, we propose VisualFactChecker (VFC), a flexible training-free pipeline designed to produce accurate and comprehensive captions for both 2D images and 3D objects. 
Fig.~\ref{fig:teaser} shows examples of captions generated by VFC and their comparisons with captions generated by GPT-4V~\citep{openai2023gpt4} and Cap3D~\citep{luo2023cap3d}.
Captions generated by VFC are faithful textural representations of the visual contents. This can also be verified by reconstructing images and 3d objects from captions using state-of-the-art text-to-image and text-to-3d models, as shown in Fig.~\ref{fig:image-recon}.

VFC focuses on tackling hallucinations and insufficient details in generated captions and is structured around three core components: \textbf{Proposer}, serving as the system's ``eye'', creating detailed caption proposals as preliminary captions by using image-to-text captioning models; 
\textbf{Large Language Model}, acting as the ``brain'', calling and summarizing information from other components, and leveraging its advanced generalization capabilities to steer the captioning process following specified captioning instructions;
\textbf{Detector and VQA models}, functioning as ``tools'' utilized by the LLM for fact-checking caption proposals, ensuring the fidelity of the final generated caption. VFC is versatile and effectively handles captioning for both 2D images and 3D objects through a unified pipeline.
Fig.~\ref{fig:vfc} shows an overview of the pipeline. The details of each component and their interplay are explained in Sec.~\ref{sec:VisualFactChecker}.

To comprehensively evaluate the generated captions, other than leveraging the commonly used CLIP-Score that primarily gauges the image-caption similarity, we propose a new metric: the CLIP-Image-Score. This metric assesses the similarity between the input image and a reconstructed image created by a text-to-image model from the caption, offering a complementary measure. 
Furthermore, we conducted a human study on Amazon Mechanical Turk for caption evaluation. 
Finally, we also performed a fine-grained evaluation by asking GPT-4V to compare and judge captions with detailed reasoning. 
The combination of CLIP-Score, CLIP-Image-Score, GPT-4V, and human study provides a more robust evaluation of captions.

We summarize our main contributions as follows:
1) We propose VisualFactChecker (VFC), a training-free pipeline to generate high-fidelity detailed 2D and 3D captions, effectively mitigating the challenge of hallucination in long captions.
(2) CLIP-Image-Score: A novel caption evaluation metric that measures the similarity between the input image and a reconstructed image from the caption.
(3) Our evaluation shows that VisualFactChecker achieves state-of-the-art results in 2D and 3D captioning tasks compared with open-sourced models.
(4) Our work shows that using an LLM to chain open-source models can achieve captioning capability on par with proprietary models such as GPT-4V.

\section{Related Work}
\label{sec:related_work}

\subsection{Image Captioning}

Image captioning has made significant progress with the advent of deep learning. Pioneering works~\citep{anderson2018bottom, donahue2015long, huang2019attention} primarily focus on integrating deep neural networks for enhanced image understanding and language generation.

Recent strides have been made with the introduction of Multimodal-Large Language Models (MM-LLMs), which are trained on extensive vision and language data. 
The general approach involves leveraging a pre-trained large language model (LLM) and a vision encoder with a projector to align with the LLM's embeddings, thus enhancing visual understanding. 
Several models have emerged as significant contributors in this domain. BLIP~\citep{li2022blip}, BLIP-2~\citep{li2023blip}, OFA~\citep{wang2022ofa}, Flamingo~\citep{alayrac2022flamingo}, Kosmos-2~\citep{peng2023kosmos}, MiniGPT-4~\citep{zhu2023minigpt}, InstructBLIP~\citep{instructblip}, LLaVA~\citep{liu2023visual, liu2023improved} have demonstrated impressive performance in single-view image captioning tasks. 
However, they exhibit varying limitations. For instance, BLIP-2 and OFA often generate overly concise captions, while others, like InstructBLIP, can produce detailed captions that often include inaccurate or hallucinatory content.
Our method aims to address these limitations by combining different models into a pipeline via an LLM, striking a better balance between accuracy and detailedness in generated captions while mitigating hallucinations.

\subsection{Large Language Models for Captioning}
Recent advancements in large language models (LLMs) like GPT-3~\citep{brown2020language}, LAMDA~\citep{thoppilan2022lamda}, PALM~\citep{chowdhery2022palm}, Llama~\citep{touvron2023llama}, GPT-4~\citep{openai2023gpt4} have demonstrated exceptional zero-shot capabilities in language analysis and summarization tasks. This proficiency has naturally extended to the multimodal domain, particularly in image-language contexts, where LLMs can summarize multimodal information in a zero-shot manner. 

Vision-blind LLMs are prominent in multimodal applications, often utilizing language-only prefixes generated by pre-trained tools. Clipcap~\citep{mokady2021clipcap} demonstrates this by using a continuous embedding as a prompt for a GPT-style language model, achieving notable performance in single-viewpoint image captioning. Similarly, Promptcap~\citep{hu2022promptcap} and PNP-VQA~\citep{tiong2022plug} leverage natural language prompts with GPT models to excel in visual question answering.

Recent methods have employed LLMs to generate image captions by summarizing initial captions or keywords from Vision-Language models.
For instance, Socratic models~\citep{zeng2022socratic} use a CLIP-based model to extract key tags from images, followed by GPT-3 with specialized prompts to create stylized captions. ChatCaptioner~\citep{zhu2023chatgpt} builds upon this by integrating ChatGPT and BLIP-2~\citep{li2023blip} in a conversational approach for question-answering about the image, and summarizing them into a caption. 
Visual Clues~\citep{xie2022visual} uses similar tags to generate a paragraph-caption. 
IC3~\citep{chan2023ic} and LLM-Fusion~\citep{bianco2023improving} use LLMs to summarize captions from existing models augmented with temperature-based sampling. 
Cap3D~\citep{luo2023scalable} extends this concept to 3D object. 

Our method differentiates itself in two critical ways: First, we focus on reducing hallucinations in captions by employing visual grounding tools, such as object detection, to fact-check captions for enhanced accuracy. Second, our pipeline can be used for captioning both 2D images and 3D objects. Unlike previous methods that rely on a single captioning model, we integrate multiple captioning sources from different models, ensuring a more comprehensive coverage of visual content to generate captions.

\begin{figure*}[t]
\centering
\includegraphics[width=\textwidth]{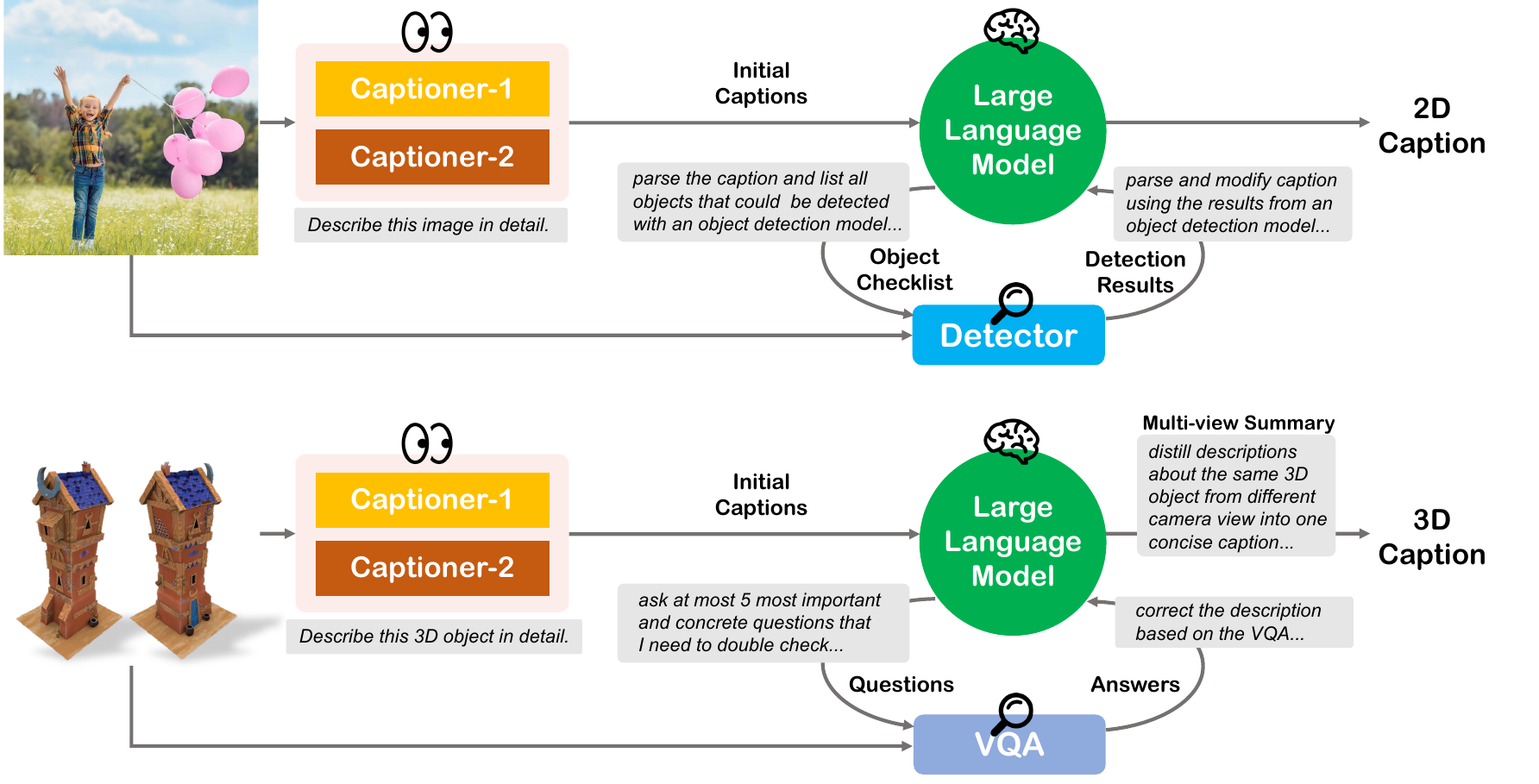}
\caption{Pipeline of the VisualFactChecker for captioning 2D images (top) and 3D objects (bottom). The process begins with the input being captioned by two multimodal captioning models (Captioner-1 and Captioner-2) to generate preliminary captions. These captions are then verified using a Large Language Model (LLM) to call object detection (Detector) and VQA models for fact-checking the captions. Finally, the LLM incorporates all the results and summarizes the final caption by following instructions.}
\label{fig:vfc}
\end{figure*}

\subsection{Hallucination in MM-LLM}

There are two popular topics on the hallucination of MM-LLMs.
(1) Hallucination evaluation: Detection approaches such as Gunjal~\etal~\citep{gunjal2023detecting} train classification models to identify hallucination. They focus on distinguishing between accurate and hallucinated content. Ground truth comparison methods~\citep{li2023evaluating, wang2023evaluation} compare model outputs with ground truth data to detect hallucinations. These techniques assess the alignment of generated captions with actual image content.
(2) Mitigation Strategies~\citep{lu2023evaluation}: Data optimization methods such as Liu~\etal~\citep{liu2023aligning} address hallucination by creating negative instances in training datasets to reduce model overconfidence. Iterative generation methods such as Wang~\etal~\citep{wang2023vigc} adopt an iterative process for caption generation, where brief answers are generated in succession and amalgamated, aiming to improve accuracy and relevance.

Our VisualFactChecker is a training-free pipeline mitigating hallucination in image captioning. Our method utilizes visual grounding tools for improved accuracy, thereby actively reducing the hallucination and offering high-fidelity captions for both 2D images and 3D objects.

\section{Visual Fact Checker}
\label{sec:VisualFactChecker}

This section introduces the key components of VisualFactChecker as shown in Fig.~\ref{fig:vfc} in detail and explains their interplay in generating accurate and detailed captions.
The following sections delve into specifics. First, we detail the pipeline for 2D image captioning (Sec.~\ref{sec:2d}), with Fig.~\ref{fig:vfc} (top) illustrating this process. Then, we explore how our approach is adapted for 3D object captioning as shown in Fig.~\ref{fig:vfc} (bottom), underscoring both shared methodologies and unique aspects relevant to 3D contexts (Sec.~\ref{sec:3d}).

\subsection{2D Image Captioning}
\label{sec:2d}

The caption generation takes three steps: 1) proposal, 2) verification, and 3) captioning. Each step is detailed below.

\noindent{\bf Proposal}: The Proposal step serves as the cornerstone of the captioning process that generates initial captions. This is achieved through the utilization of advanced image-to-text models, specifically ``LLaVA'' and ``Kosmos2''. These models are trained on expansive datasets, enabling them to comprehend and interpret visual content effectively. By analyzing the input image, they suggest various preliminary captions, each reflecting different facets and interpretations of the image (Fig.~\ref{fig:vfc} top). The rationale behind using multiple image-to-text multimodal LLMs lies in the complexity of adequately capturing an image's essence in a single attempt. Since an image can be accurately described in numerous ways, different models bring unique perspectives, thereby encompassing a broader range of information present in the image. Although the initial captions proposed may not possess perfect fidelity, the primary objective at this stage is to generate captions that are as comprehensive as possible. Fig.~\ref{fig:vfc} displays the specific prompts we used for each step,  with more details in Appendix~\ref{appendix:models_prompts}.

\noindent{\bf Verification and Captioning}: The goal of the verification step is to scrutinize and rectify any inaccuracies or hallucinations in the captions during the proposal step. It employs a combination of a Large Language Model (LLM) and grounding tools, including an open-vocabulary object detection model and/or a visual question answering (VQA) model. Here the LLM can be GPT-4 or Llama2. As shown in Fig.~\ref{fig:vfc} (top), the process involves the following steps: 
Step 1: LLM first summarizes the initial detailed descriptions from different MM-LLMs into a single, detailed caption. While this caption is comprehensive, it may not always be accurate.
Step 2: The LLM then analyzes this synthesized caption, identifying all objects that could be verified by object detection and summarizing an object checklist. In 2D image captioning, the focus is on eliminating hallucinations, particularly descriptions of non-existent objects in the image. Identifying these objects is crucial for the subsequent verification process.
Step 3: Taking the object checklist as input, an open-vocabulary object detection model examines candidate objects in the checklist and determines their presence in the image. This step is pivotal in validating the existence of objects mentioned in the caption, thus supporting the fidelity of the caption.

After verification, we go to the last captioning step: Based on the object detection results, the LLM revises the summarized single detailed caption. Each object described in the caption is cross-checked; if detected, it remains unchanged, while undetected objects are considered potential hallucinations and are removed from the caption. This step results in a final caption that is both detailed and reliable. The underlying assumption is that the detection model, serving as an object grounding expert, provides more reliable results than a general image descriptor.

In the verification and captioning steps, the LLM plays a pivotal role as a ``brain''. It starts by parsing the initial caption and identifying key objects for detailed examination. The LLM then meticulously assesses whether each object mentioned actually appears in the image based on detection results. Following this thorough analysis, it refines and revises the initial captions, transforming them into final versions that are both coherent and richly detailed. 
The LLM is instrumental in guaranteeing linguistic fluency, ensuring that the captions not only accurately represent the image but also maintain the necessary level of detail for high-fidelity captioning.
Moreover, the LLM can follow complex instructions to write the captions in a specified style, such as a caption that only mentions the foreground objects without mentioning the background.
Fig.~\ref{fig:vfc} displays the specific prompts used for each step.

\subsection{3D Object Captioning}
\label{sec:3d}
The 3D object captioning process follows a similar structural pipeline to that of 2D images, with a few key distinctions in certain steps, as depicted in Fig.~\ref{fig:vfc} (bottom). In 3D captioning, an object may present multiple views, each offering unique information. The comprehensive caption for a 3D object is derived by integrating the perspectives from all these views. For each view, VisualFactChecker is employed to create a detailed, high-fidelity description. Subsequently, the LLM (GPT-4 or Llama-2) is used to amalgamate the information from all views, producing a unified caption for the 3D object.
In particular, for each view's captioning, we have the same three-step approach akin to 2D image captioning. In the proposal step, LLaVA-1.5 and InstructBLIP are utilized for generating initial detailed descriptions. We opt out of using Kosmos2 for single 3D objects due to its less effective performance in providing detailed descriptions, possibly linked to its reliance on an implicit detection model. Additionally, a slightly modified prompt is used (see Fig.~\ref{fig:vfc} bottom), which incorporates 3D-specific considerations.
In the verification and captioning step, we primarily address hallucinations related to the attributes of 3D objects, such as shape and color. To mitigate these inaccuracies, rather than enumerating potential objects, we employ the LLM to generate five critical questions that could influence a text-to-3D generation model in reconstructing the 3D model. Following this, we utilize VQA models (specifically LLaVA-1.5) to respond to these questions based on the input 3D object view image. Subsequently, the LLM amends the initial caption in accordance with the answers provided by the VQA model. We operate under the assumption that answering targeted questions results in fewer hallucinations compared to generating a general description.
Once the caption for each individual view is complete, the LLM synthesizes these multiple perspectives into a singular, comprehensive caption for the entire 3D object. The prompts used for the LLM at each stage are detailed in Appendix~\ref{appendix:models_prompts}.

\section{CLIP-Image-Score}
\label{sec:clip-image-score}
\begin{figure}
\centering
\includegraphics[width=\linewidth]{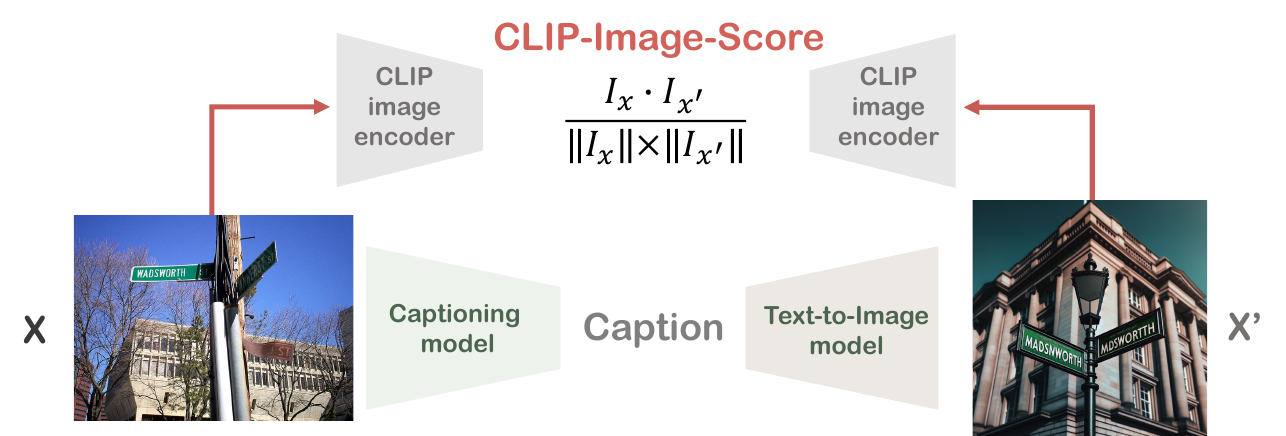}
\caption{The CLIP-Image-Score pipeline evaluates caption accuracy by encoding an original image \( X \) into a feature representation \( I_X \) using a CLIP image encoder. A captioning model generates a caption that is then input into a text-to-image model to reconstruct an image \( X' \), which is encoded to \( I_{X'} \). The score is computed by assessing the cosine similarity between \( I_X \) and \( I_{X'} \), providing a measure of the caption's fidelity and hallucination detection.}
\label{fig:clip-image-score}
\end{figure}

Accurate evaluation of caption correctness and detailedness is paramount in determining the performance of an image captioning model. Traditional metrics like the CLIP-Score~\citep{hessel2021clipscore} have served as a standard for measuring the alignment between generated captions and their corresponding images. 
However, our CLIP-score may lack the sensitivity needed to detect the specific issue of hallucination within captions. 

We present the CLIP-Image-Score, an alternative metric specifically developed to reflect the subtleties of caption quality. This metric is different from CLIP-Score by introducing an additional reconstruction step. Specifically, the CLIP-Image-score evaluates the similarity between the original image and a reconstructed version of the image generated by a fixed text-to-image model using the caption as a prompt. By comparing the raw image to its reconstructed image, the metric is able to detect discrepancies indicative of hallucination, thus providing a different perspective of the caption quality assessment. 
The underlying principle of the CLIP-Image-Score is the recognition that multiple ``correct'' captions may exist for a single image. However, it's only when a caption is both ``detail'' and ``correct'' that the reconstructed image closely resembles the original. 
Moreover, any hallucinations present in the caption become evident in the reconstructed image. Fig.~\ref{fig:image-recon} presents examples of such reconstructions. For instance, consider the results from LLaVA-1.5 shown in the third column. The caption generated for the first image falsely mentions ``several other people in the background''. This error is clearly reflected in the image reconstructed by the text-to-image generator. In essence, comparing the two images indirectly ensures alignment between the image and its caption, thereby providing a complementary method to assess the quality of the caption than directly comparing the image and caption.

The CLIP-Image-Score evaluation process is depicted in the following steps:
\begin{itemize}
    \item \textbf{Caption Generation}: An original image \( X \) is input into a captioning model, which generates a caption.
    
    \item \textbf{Caption-to-Image Reconstruction}: This generated caption is then used as input for a text-to-image model, which creates a reconstructed image \( X' \) that visually represents the textual description.

    \item \textbf{Raw Image Encoding}: The original image \( X \) is processed through a CLIP image encoder, translating the visual content into an encoded representation \( I_X \).
    
    \item \textbf{Reconstructed Image Encoding}: The reconstructed image is also processed through the CLIP image encoder to obtain its encoded representation \( I_{X'} \).
    
    \item \textbf{Score Calculation}: Finally, the encoded representations of the original and reconstructed images are compared to calculate the CLIP-Image-Score. The score is given by the cosine similarity, which assesses the congruence between \( I_X \) and \( I_{X'} \):
    \begin{equation}
        \text{CLIP-Image-Score} = \frac{I_X \cdot I_{X'}}{\|I_X\| \times \|I_{X'}\|}
    \end{equation}
\end{itemize}

Most notably, CLIP-Image-Score offers a sensitive measure for detecting hallucinations. In scenarios where the generated caption includes elements that are not in the original image, the reconstructed image will also likely contain these discrepancies. By comparing the original and reconstructed images, the CLIP-Image-Score can effectively highlight these differences, offering a clearer insight into the fidelity and accuracy of the generated caption.

Furthermore, CLIP-Image-Score turns a cross-modality comparison into a more intuitive comparison in the same image modality (as shown in Fig.~\ref{fig:clip-image-score}). 
CLIP-Image-Score represents a new complementary perspective for image captioning evaluation. By leveraging the capabilities of text-to-image models and focusing on the congruence between the original and reconstructed images, it provides an accurate assessment of caption quality, particularly in identifying and measuring hallucinations, thereby enhancing the overall reliability of caption generation systems.













\section{Experiments}
\label{sec:exp}



This section presents a thorough evaluation of captioning models across both 2D and 3D visual content, employing a variety of datasets and methodologies. Table~\ref{table:compare-summary} provides a summary of our comprehensive evaluation experiments.

\begin{table}[t]
    \centering
    \footnotesize
\resizebox{\linewidth}{!}{
\begin{tabular}{cccc}
\toprule
\textbf{Eval} & \textbf{Input pairs for evaluation} & \textbf{Method} & \textbf{Reference}\\
\hline
\multirow{3}{*}{2D} & \multirow{3}{*}{Raw image  \ \ \ \  Caption} & CLIP-Score & Table~\ref{table:2d-clip-score}\\

&    & Human evaluation & Fig.~\ref{fig:AMT}\\
&    & CPT4V evaluation & Fig.~\ref{fig:gpt4}\\

\cline{2-3}
&  Raw image \ \ \ \ Image(recon) & CLIP-Image-Score & Table~\ref{table:2d-clip-score}\\

\hline

\multirow{3}{*}{3D} & \multirow{2}{*}{Multi-view (raw)  \ \ \ \  Caption} & CLIP-Score &  Table~\ref{tab:3d-clip-score}\\

&   & GPT4V evaluation & Fig.~\ref{fig:gpt4}\\

\cline{2-3}
& Multi-view (raw)  \ \ \ \   Multi-view (recon) & CLIP-Image-Score& Table~\ref{tab:3d-clip-score}\\

\bottomrule
\end{tabular}
}
\caption{Summary of evaluation methods and results.}
\label{table:compare-summary}
\end{table}








\subsection{Overall: CLIP-Score and CLIP-Image-Score}

\noindent{\bf 2D image captioning.} 
{\bf Dataset}: Our evaluation utilized 5,000 COCO test images from the Karpathy split.
{\bf Baseline methods}: We benchmarked against state-of-the-art captioning models, including BLIP-2~\citep{li2023blip}, InstructBLIP~\citep{instructblip}, and LLaVA-1.5~\citep{liu2023improved}. The evaluation focused on each model's ability to produce accurate, detailed, and coherent captions that effectively encapsulate the essence of the images.
{\bf Evaluation Metric}: We employed two metrics: CLIP-Score \citep{hessel2021clipscore} and CLIP-Image-Score (Sec.~\ref{sec:clip-image-score}).
The CLIP-Score, a prevalent metric in image caption quality assessment, involves processing the raw image through the CLIP image encoder and the caption through the CLIP text encoder. The resultant embeddings are then compared for cosine similarity, with a higher score indicating greater semantic resemblance between the image and the caption.
For our analysis, we first calculated the CLIP-Score for each image-caption pair, then averaged these scores across all 50,000 text/image pairs, scaling the result by a factor of 100.
Table~\ref{table:2d-clip-score} displays the comparative performance of various image captioning methods on the 5,000 COCO test set images. The results demonstrate that our VisualFactChecker surpasses all baseline methods in performance.




\begin{table}[h]
\centering
\small
\resizebox{\linewidth}{!}{
\begin{tabular}{lll}
\toprule
\textbf{Captioning Method} & \textbf{CLIP-Score (\%) $\uparrow$} & \textbf{CLIP-Image-Score (\%) $\uparrow$} \\
\hline
Human Label (COCO GT) & 30.36~~\textcolor{red}{(-2.54)} & 71.21~~\textcolor{red}{(-2.40)}\\
BLIP2 & 30.11~~\textcolor{red}{(-2.79)} & 70.79~~\textcolor{red}{(-2.82)}\\
InstructBLIP & 31.45~~\textcolor{red}{(-1.45)} & 72.95~~\textcolor{red}{(-0.66)}\\
LLaVA-1.5 & 32.08~~\textcolor{red}{(-0.82)} & 73.24~~\textcolor{red}{(-0.37)}\\
Kosmos-2 & 32.32~~\textcolor{red}{(-0.58)} & 73.28~~\textcolor{red}{(-0.33)}\\
\hline
VisualFactChecker (Ours) & \textbf{32.90} & \textbf{73.61}\\
\hline
\end{tabular}
}
\caption{Image captioning comparison with different metrics on 5000 COCO test set in Karpathy split, we use raw image and caption as input pairs for evaluation.}
\label{table:2d-clip-score}
\end{table}

As outlined in Sec.~\ref{sec:clip-image-score}, the CLIP-Image-Score provides a complementary view to assess the quality of image captions. This metric is derived by comparing the cosine similarity between the CLIP embeddings of two images: the original image and a reconstructed image, which is generated using the provided caption through a text-to-image generation model. A higher CLIP-Image-Score signifies a more accurate and effective image caption. For this process, Stable Diffusion XL (SDXL)~\citep{podell2023sdxl} is utilized as the designated text-to-image model to reconstruct images based on the generated captions.
Table~\ref{table:2d-clip-score} presents the CLIP-Image-Scores obtained for the 5000 images in the COCO test set, where our method outperforms all baseline methods. 




\noindent{\bf 3D object captioning.} 
{\bf Dataset}: 1,000 3D objects sampled from Objaverse dataset \citep{deitke2023objaverse}.
{\bf Baseline methods}: We use state-of-the-art 3D object captioning model Cap3D \citep{luo2023cap3d} as the baseline. Cap3D uses 8 view images to generate the final object caption, our VisualFactChecker uses only 2 views to generate the object caption.
{\bf Evaluation Metric}: CLIP-Score and CLIP-Image-Score on multiple views rendered from 3D objects.
To evaluate the similarity of a 3D object and the generated caption, we evaluate the similarity of the caption with the multi-view images used to generate the caption. Specifically, we evaluate the similarity of the generated caption with the two views that were used to generate the caption and use the average score to represent the CLIP-Score. 
Table.~\ref{tab:3d-clip-score} shows the performance of 3D object captioning methods on 1,000 3D objects from Objaverse dataset. VisualFactChecker outperforms Cap3D.




\begin{table}[t]
\centering
\small
\resizebox{\linewidth}{!}{
\begin{tabular}{lll}
\toprule
\textbf{Captioning Method} & \textbf{CLIP-Score (\%) $\uparrow$} & \textbf{CLIP-Image-Score (\%) $\uparrow$} \\
\hline
Cap3D & 33.44~~\textcolor{red}{(-0.57)} & 79.88~~\textcolor{red}{(-0.44)} \\
\hline
VisualFactChecker (Ours) & \textbf{34.01} & \textbf{80.32} \\
\hline
\end{tabular}
}
\caption{3D object captioning comparison with different metrics on 1000 objects in Objaverse. For CLIP-Score, we use the average score of two views for evaluation. For CLIP-Image-Score, we use an off-the-shelf text-to-3D model, MVDream, to generate 3D models from 3D captions. We compare two views of the raw object and the same views of generated 3D object for evaluation.}
\label{tab:3d-clip-score}
\end{table}

We also use CLIP-Image-Score to evaluate the 3D caption quality. CLIP-Image-Score needs reconstructed images to compare with the raw images. We treat the two views that were used to generate the 3D object caption as the raw image. To obtain the reconstructed image, we use an off-the-shelf text-to-3D generation model, MVDream, to generate a 3D object given the generated 3D object caption. We then render the same two views of images based on the generated 3D object, and we calculate the CLIP-Image-Score between the raw image and the rendered image.
Table.~\ref{tab:3d-clip-score} shows the CLIP-Image-Score on 1000 objects in Objaverse dataset.










\begin{figure}[h]
\centering
\includegraphics[width=\linewidth]{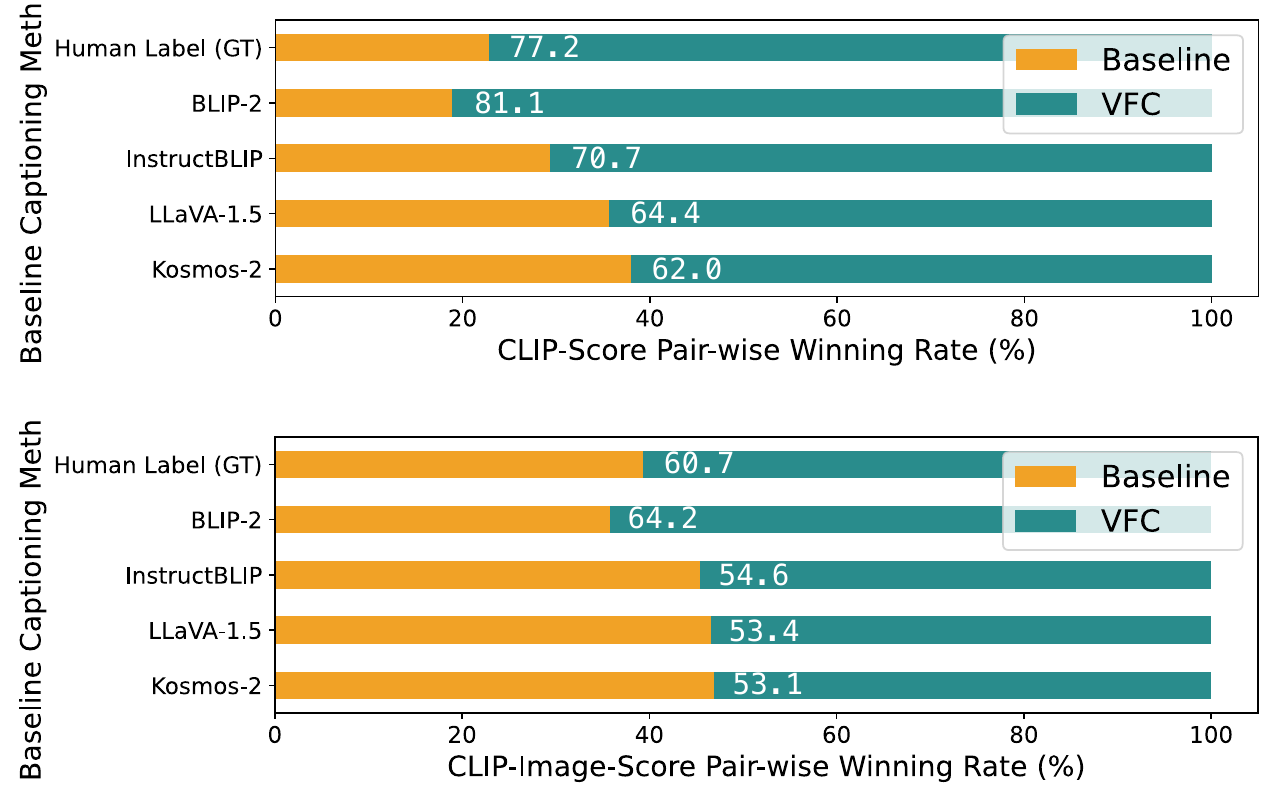}
\caption{2D image captioning comparison with pair-wise winning rate. VisualFactChecker (\textcolor{teal}{VFC}) outperforms all \textcolor{orange}{baseline methods} on both CLIP-Score (top) and CLIP-Image-Score (bottom).}
\label{fig:image-wise}
\end{figure}
\subsection{Per Image Evaluation: Wining Rate}
CLIP-Score and CLIP-Image-Score indicate an overall performance comparison, which shows an average score among all 5000 images. The average score may be dominated by a small group of images that have extremely high or low scores. To zoom in and show a more detailed comparison, we try to answer the following question: Given an image, what is the probability that one method performs better than another method on caption generation? To answer this question, we need to go over each image and calculate the winning rate for a pair of methods.

Specifically, for each image, we compare the CLIP-Score of our VisualFactChecker caption against the captions generated from different baselines respectively, and calculate the
wining probability of our method and the baselines. 
Fig.~\ref{fig:image-wise} shows the results, for example, we can see that in the pair-wise comparison, our VisualFactChecker performs better (higher CLIP-Score) than LLaVA-1.5 on 64.4\% of 5000 images (3220 images).

Calculating the winning rate over all images provides a more detailed analysis that zooms in on the comparison of each image, which shows a complementary view than overall average CLIP-Score.


\subsection{Fine-grained Evaluation: Human and GPT-4V}

The CLIP-Score and CLIP-Image-Score offer a general comparison of overall performance. A pairwise per-image winning rate provides a more specific analysis, evaluating performance on individual images. 
However, the research highlighted in related studies \citep{kotar2023these} indicates that the CLIP-Score may not be ideally suited for image-to-image comparison tasks. Furthermore, relying on a single score fails to provide a nuanced comparison across criteria, such as accuracy and level of detail. We use Human evaluation and GPT-4V to provide a more fine-grained evaluation.

\begin{figure}[h]
\centering
\includegraphics[width=\linewidth]{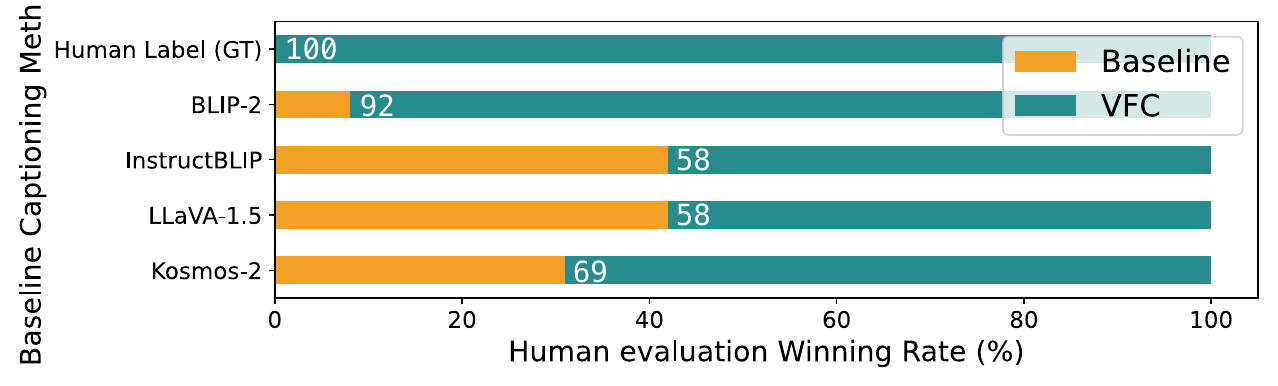}
\caption{Amazon Mechanical Turk human evaluation results.}
\label{fig:AMT}
\end{figure}

\noindent{\bf Human evaluation using Amazon Mechanical Turk (AMT).}
We employed a pairwise comparison strategy. From the COCO dataset, we randomly selected 100 images out of 5000. 
For each image, our caption was compared against 5 baseline captions respectively. 
To reduce variance, each comparison was done by 3 different AMT workers and we used their majority voting as the final selection. 
This resulted in a total of $1500$ comparisons collected on AMT.
AMT UI is shown in the appendix. 
The workers were presented with two competing captions — one from a baseline method and one from our VisualFactChecker, in randomized order. They were instructed to select the better caption describing the image based on 3 aspects: correctness, detailness, and fluency. 
Results in Fig.~\ref{fig:AMT} show our captions are more preferred by humans.
The human evaluation instruction and web UI is shown in Appendix~\ref{appendix:eval}.


\noindent{\bf GPT-4V evaluation.} Our study applied GPT-4V for evaluating captions in a manner akin to the caption evaluation process used in DALLE-3. We use the same randomly selected 100 images from COCO as in Human evaluation.
For each image, we considered the captions generated by 5 baseline methods alongside the caption produced by our VisualFactChecker. We then presented GPT-4V with the raw image, our reference caption, and the four baseline captions. Our designed prompt instructed GPT-4V to compare each baseline caption against our reference caption, focusing on two primary aspects: correctness and detail. 
GPT-4V was tasked with providing a pairwise, detailed comparison for each pair, including justifications for its assessments. Based on these comparative insights, GPT-4V classified each baseline method caption as either ``better'' or ``worse'' than our VisualFactChecker. Fig.~\ref{fig:image-wise} shows the comprehensive results. 
More details about the GPT-4V evaluation prompt and examples are shown in Appendix~\ref{appendix:eval}.







\begin{figure}[h]
\centering
\includegraphics[width=\linewidth]{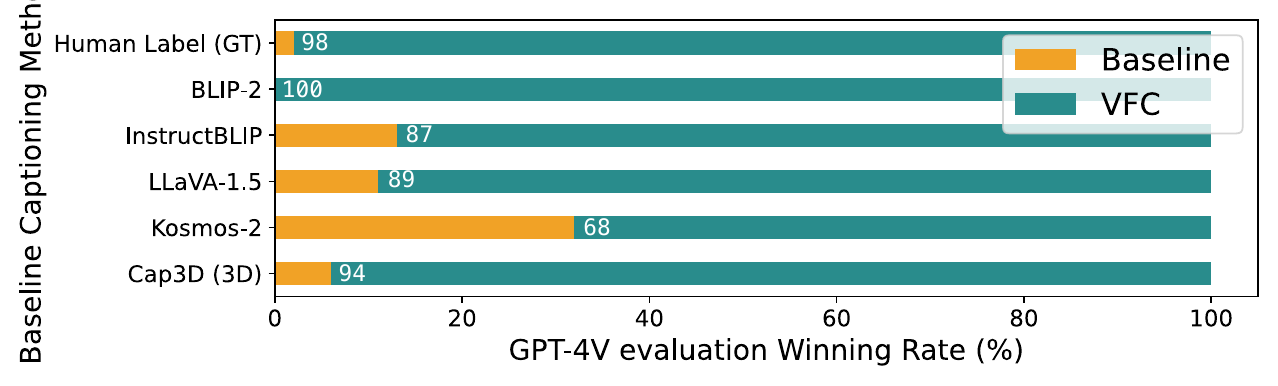}
\caption{GPT-4V evaluation results. Our captions are significantly better than baselines.}
\label{fig:gpt4}
\end{figure}


\subsection{Ablation Study}

In our ablation study, we explore the impact of various components on performance. For 2D captioning tasks, we assess the efficacy of initial captioning models, LLaVA-1.5 and Kosmos-2, using the CLIP-Score metric for the captions they generate on the same 5000 COCO test images. Additionally, we ablate our method's performance in the absence of the verification (fact checker) step, which aims to mitigate hallucinations through detection grounding. Table~\ref{table:ablation} shows the detailed results. 
Likewise, in the context of 3D object captioning, we evaluate the individual contributions of initial captioners, namely LLaVA-1.5 and InstructBLIP on the same 1000 Objaverse 3D objects. We further investigate the performance of our methodology without the fact checker, which in this case operates by leveraging a VQA model to reduce hallucinations. Table~\ref{table:ablation} shows the detailed results.
These results highlight the significance of fact checker in our approach.


\begin{table}
    \centering
    \small
\resizebox{\linewidth}{!}{
\begin{tabular}{ccl}
\toprule
\multicolumn{2}{c}{\textbf{Methods or Steps} } & \textbf{CLIP-Score} \\
\hline
\multirow{4}{*}{2D} & LLaVA-1.5 & 32.08 \textcolor{red}{(-0.33)} \\
& Kosmos-2 & 32.32  \textcolor{red}{(-0.09)} \\
& VisualFactChecker (w/o fact check) & 32.41  \\
& VisualFactChecker  & \textbf{32.90 \textcolor{blue}{(+0.49)}}  \\
\hline
\multirow{5}{*}{3D} & LLaVA-1.5 & 32.05 \textcolor{red}{(-0.66)} \\
& InstructBLIP & 32.51 \textcolor{red}{(-0.20)} \\
& VisualFactChecker (w/o fact check) & 32.71 \\
& VisualFactChecker & \textbf{34.01 \textcolor{blue}{(+1.30)}} \\
\bottomrule
\end{tabular}
}
\caption{Ablation study on captioning 2D images (5000 COCO test dataset) and 3D objects (1000 Objaverse).}
\label{table:ablation}
\end{table}

\subsection{Qualitative Results and Prompt Following}
Other than quantitative evaluation results, we show more qualitative examples of VisualFactChecker for 2D and 3D captions in Appendix~\ref{appendix:vis}.

By leveraging an LLM, VisualFactChecker can follow complex instructions to write captions in various styles. Examples are shown in Appendix~\ref{appendix:rewrite}.

\section{Conclusion}
We propose the VisualFactChecker (VFC), a training-free pipeline to generate high-fidelity and detailed captions. By utilizing an LLM to chain multimodal models and object detection and VQA models, VFC reduces hallucination in long captions. We conducted a comprehensive caption evaluation using different metrics, including 1) image-text similarity using CLIP-Score, 2) image-reconstructed image similarity using our proposed CLIP-Image-Score, 3) human study, and 4) fine-grained evaluation using GPT-4V. Compared with open-sourced captioning models, our method achieves state-of-the-art in both 2D and 3D captioning.
Our work shows combining open-sourced models into a pipeline can significantly close the captioning performance gap with proprietary models like GPT-4V.
In the future, we plan to improve our pipeline further by including more components for fact-checking and making it more automatic in deciding which components to use.

\noindent{\bf Acknowledgments} We would like to thank Siddharth Gururani for helping with our human evaluation using Amazon Mechanical Turk; Haochen Wang for his help in pre-processing 3D data. We also thank Qinsheng Zhang, Yogesh Balaji, and Yen-Chen Lin for their helpful discussion.

\clearpage

{
    \small
    \bibliographystyle{ieeenat_fullname}
    \bibliography{main}
}

\appendix
\clearpage
\maketitlesupplementary

\section{More Details on Models and Prompts}
\label{appendix:models_prompts}
The models used in VisualFactChecker and baselines are:
\begin{itemize}
    \item \textbf{Caption Proposer:}
    \textit{\href{https://huggingface.co/Salesforce/blip2-opt-2.7b}{BLIP-2-OPT-2.7B}}, \textit{\href{https://github.com/salesforce/LAVIS/tree/main/projects/instructblip}{InstructBLIP-7B}}, 
    \textit{\href{https://huggingface.co/liuhaotian/llava-v1.5-13b}{LLaVA-1.5-13B}}, 
    \textit{\href{https://huggingface.co/ydshieh/kosmos-2-patch14-224}{Kosmos-2}}
    \item \textbf{LLMs:}
    \textit{\href{https://platform.openai.com/docs/models/gpt-4-and-gpt-4-turbo}{GPT-4-0613}}, \textit{\href{https://huggingface.co/meta-llama/Llama-2-70b-chat-hf}{Llama-2-70B-chat}}
    \item \textbf{Detector:}
    \href{https://github.com/IDEA-Research/Grounded-Segment-Anything}{Grounding DINO}
    \item \textbf{VQA:}
    \textit{\href{https://huggingface.co/liuhaotian/llava-v1.5-13b}{LLaVA-1.5-13B}}
\end{itemize}

Below are the prompts we used in VisualFactChecker for captioning 2d images and 3d objects.

\begin{tcolorbox}[colback=mplblue!3,colframe=mplblue!75!white,title=\textsc{VisualFactChecker prompts (2D images)},left=0.2ex,right=0.2ex,top=0.2ex,bottom=0.2ex]
\fontsize{8}{8}\selectfont
\emph{\textbf{Proposal (LLaVA-1.5 / Kosmos-2):}} \\
Describe this image in detail.\\

\emph{\textbf{Verification step-1 (GPT-4 / Llama-2):}}\\
This is a hard problem. Carefully summarize in ONE detailed caption based on the following two captions by different (possibly incorrect) people describing the same scene. Be sure to describe everything, and avoid hallucination. \\

\emph{\textbf{Verification step-2 (GPT-4 / Llama-2):}}\\
I want to use an object detector to check the correctness of an image caption obtained by an image caption model. Can you help to parse the caption below and list all objects that could be detected with an object detection model in the image? Please only list the object name and ignore the description. Please use singular for all listed objects.
\\
Caption: \{\}.

Please concatenate them together with ``. '' as separation.\\

\emph{\textbf{Verification step-3 (Grounding DINO):}}\\
N/A (Grounding DINO examines candidate objects in the checklist above and determines their presence in the image.)\\

\emph{\textbf{Captioning(GPT-4 / Llama-2):}}\\
Objective: parse and modify image captions using the results from an object detection model (may have hallucination).
\\\\
I will put the detection results to you in the following format: [``object'': detected object name, ``number'': number of detected object (N)]. Please follow the following steps:
\\\\
Instructions:
\\
Parse the object in the caption, (Note: only parse and modify the object (not color, action, size, shape, or other descriptions))\\
1. If the object was detected by the detection model, keep everything including all descriptions. For instance, if the original caption is: ``a black and white panda toy'', if the toy was detected, keep all content even though the ``panda'' and ``black and white'' are not detected. Keep all descriptions about color, shape, actions .etc.\\
2. If the subject object was not detected, remove only the object. Do NOT remove color, shape, action, text and other descriptions.\\
3. Only decrease the object number if the detected object number is smaller than the caption number.
\\\\
This is a hard problem. Please minimize modifications of the caption, and list all changes made along with the reasoning.
\\
---BEGIN Detection results: ---

\{\}

---END Detection results---

---BEGIN Raw caption: ---

\{\}

---END Raw caption---
\\\\
Please give the output in the following format:

Modification:

Updated caption:
\end{tcolorbox}

We find Llama-2 may encounter difficulties in the last 2D captioning step due to the complexity of the prompt. A workaround is to use a script to compare the detection results with the object list from step-2 and identify objects to be removed. Then, employ Llama-2 solely for removing these objects and summarizing the description.

\begin{tcolorbox}[colback=mplblue!3,colframe=mplblue!75!white,title=\textsc{VisualFactChecker prompts (3D objects)},left=0.2ex,right=0.2ex,top=0.2ex,bottom=0.2ex]
\fontsize{8}{8}\selectfont
\emph{\textbf{Proposal (LLaVA-1.5):}} \\
Please describe the details of the 3D object, the detailed description will be used for a text to 3d model to generate this 3D object. Please provide details of the shape, color of each part, avoid imagination and solve it step by step.\\

\emph{\textbf{Proposal (InstructBLIP):}} \\
Describe the 3D object in detail, step by step.\\

\emph{\textbf{Verification step-1 (GPT-4 / Llama-2):}}\\
This is a hard problem. Carefully summarize in ONE detailed caption based on the following two captions by different (possibly incorrect) people describing the same 3D object. The detailed caption will be used for a text to 3D model to generate this 3D object. Be sure to describe everything, and avoid hallucination.\\

\emph{\textbf{Verification step-2 (GPT-4 / Llama-2):}}\\
I have a description of a 3D object, the detailed caption will be used for a text to 3d model to generate the same 3D object. Some part of the description may have some hallucination, so I want to use a VQA model to double check some key description, Please ask at most 5 most important and concrete questions that I need to double check to improve the fidelity of the description. Please focus on the factors that influence the final text to 3D model generation.
\\\\
Raw Caption: \{\}
\\
Please output the 5 questions in a python list.\\

\emph{\textbf{Verification step-3 (LLaVA-1.5):}}\\
N/A (LLaVA-1.5 takes questions above and raw view image as input and give answers).\\

\emph{\textbf{Single view Captioning (GPT-4 / Llama-2):}}\\
I have a description of a 3D object, the detailed caption will be used for a text to 3d model to generate the same 3D object. Some part of the description may have some hallucination, so I use a VQA model to double check some key description.\\
Here is the original description that may contain hallucination:
\{\}\\
Here are the questions and answers from a VQA model:
\{\}\\\\
Please correct the description based on the VQA. I want to use the description as a prompt for a text-to-3D generation model to generate the same 3D object.\\

\emph{\textbf{Object Captioning(GPT-4 / Llama-2):}}\\
Given a set of descriptions about the same 3D object from different camera views, please distill these descriptions into one concise caption:\\
Camera View 1 description: \{\}
\\
Camera View 2 description: \{\}
\end{tcolorbox}

\section{Details on Human and GPT-4V Evaluaiton}
\label{appendix:eval}
\begin{figure*}[h]
\centering
\includegraphics[width=\linewidth]{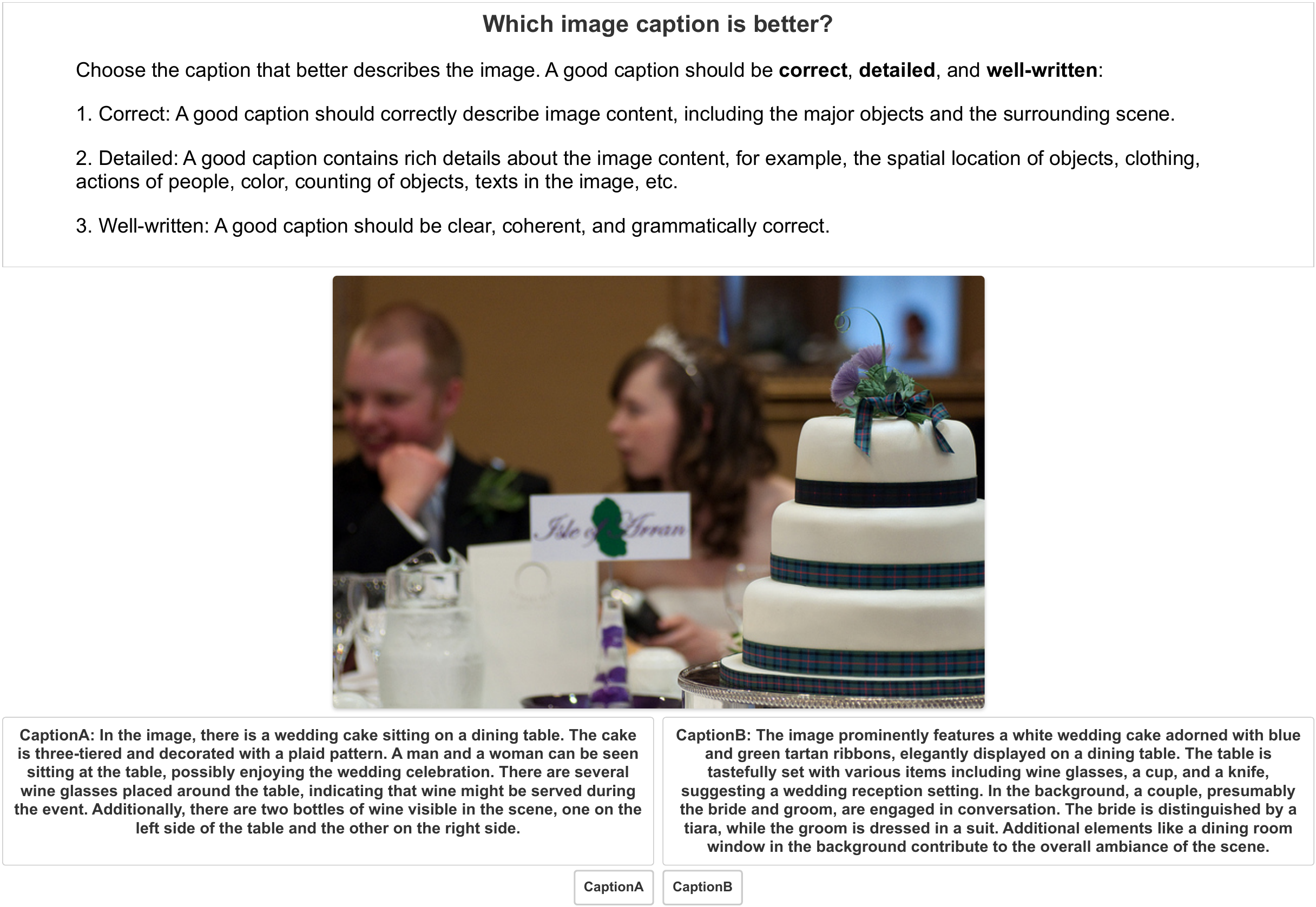}
\caption{Amazon Mechanical Turk web user interface.}
\label{fig:AMT-UI}
\end{figure*}

Fig.~\ref{fig:AMT-UI} shows the Amazon Mechanical Turk human evaluation web UI.
For GPT-4V evaluation, inspired by DALLE-3~\citep{dalle3}, we craft a single prompt for evaluating all captions for a given image using GPT-4V (\textit{\href{https://platform.openai.com/docs/guides/vision}{gpt-4-vision-preview}}). The prompt is as follows.

\begin{tcolorbox}[colback=mplblue!3,colframe=mplblue!75!white,title=\textsc{GPT-4V evaluation prompt (2D images)},left=0.2ex,right=0.2ex,top=0.2ex,bottom=0.2ex]
\fontsize{8}{8}\selectfont
You are responsible for judging the quality of image captions generated by computer programs. You will be presented with an image and captions describing the image. One of the captions is a reference caption and you are responsible for comparing other captions against the reference caption, one by one. Please evaluate captions in the following two aspects:
\\
1. Correctness. A good caption should correctly describe image content, including the major objects and the surrounding scene.
\\
2. Detailedness. A good caption should contain rich details about the image content, for example, spatial location of objects, clothing and actions of people, color, counting of objects, texts in the image, etc.
\\\\
Given the image and a reference caption, you need to make decisions as to whether or not the given captions are better or worse than the reference caption in describing the overall image. 
\\\\
This is a difficult task. You should first evaluate the reference caption for the given image. And then evaluate each caption in the same way and compare against the reference caption. You should think out loud about your eventual conclusion for each caption, enumerating reasons why each of the captions is better or worse compared with the reference caption. After thinking out loud, you should output either ``Better'' or ``Worse'' for each caption depending on whether you think the caption is a better description of the image than the reference caption.
\\
Reference Caption: \{\}\\
Caption 1: \{\}\\
Caption 2: \{\}\\
Caption 3: \{\}\\
Caption 4: \{\}\\
\\
You must keep your thinking out loud and respond in less than 500 words. Please summarize your judgment at the end using the following format:
\\
Caption 1: Better or Worse\\
...\\
Caption 4: Better or Worse
\end{tcolorbox}

We provide an example of GPT-4V's response, corresponding to Fig.~\ref{fig:supp-2d-vis-2} (d), where the reference caption is our proposed VisualFactChecker, and caption 1 to 4 are BLIP-2, InstructBLIP, Kosmos-2 and LLaVA-1.5, respectively. 

\begin{tcolorbox}[colback=mplgreen!3,colframe=mplgreen!75!white,title=\textsc{GPT-4V evaluation response (2D images)},left=0.2ex,right=0.2ex,top=0.2ex,bottom=0.2ex]
\fontsize{7}{7}\selectfont
To evaluate each caption, I will consider the aspects of correctness and detailedness against the reference caption.\\
Reference Caption: The reference caption accurately describes that a person is snowboarding and performing a jump while kite flying, mentioning the colors of the outfit and the snowboard. It also provides context for the backdrop and gives a sense of the snowboarder's abilities.\\
Caption 1: This caption introduces the term ``kite surfing'', which could be misleading because that term is typically associated with water rather than snow. It lacks the detail of the outfit and the specifics of the snowboarding action.\\
Caption 2: This caption provides a summarization close to the reference but omits the colors of the outfit and the snowboard. It does, however, accurately affirm the skill and control of the snowboarder and the backdrop of the scene.\\
Caption 3: This caption mentions objects that are not present in the image, such as other people in the scene. It retains the mention of the green outfit and emphasizes the height of the jump and the kite's position but suggests parasailing or kiteboarding, which might not be accurate terms for this activity.\\
Caption 4: The description of the snowboarder's action and outfit matches the image well, and it adds the color of the sky. It also introduces additional context by mentioning other spectators in the scene, but it does not mention the kite flying aspect, which is essential for the activity depicted.\\
Judgment:\\Caption 1: Worse\\Caption 2: Worse\\Caption 3: Worse\\Caption 4: Worse
\end{tcolorbox}

Similarly, we design prompt for evaluating captions for 3D object, given two rendered views (front view and back view). The prompt is shown as follows.

\begin{tcolorbox}[colback=mplblue!3,colframe=mplblue!75!white,title=\textsc{GPT-4V evaluation prompt (3D objects)},left=0.2ex,right=0.2ex,top=0.2ex,bottom=0.2ex]
\fontsize{8}{8}\selectfont
You are responsible for judging the quality of captions generated by computer programs designed for describing 3d models. You will be presented with two rendered images of the same 3d model and two captions describing the 3d model. Please evaluate captions in the following aspects:
\\\\
1. Correctness. A good caption should correctly describe the 3d object, including what the object is and the major properties of the object.\\
2. Detailedness. A good caption should contain rich details about the 3d object, for example, part, color, material, function, etc.
\\\\
Given rendered images of the 3d model and two captions, you need to make decisions on which caption is better in describing the overall 3d object. 
\\\\
This is a difficult task. You should first evaluate two captions based on rendered views of the 3d object. You should think out loud about your eventual conclusion for each caption, enumerating reasons why each of the captions is better or worse compared with the other. After thinking out loud, you should output either ``Caption 1'' or ``Caption 2'' depending on which caption you think is a better description of the 3d model.
\\\\
Caption 1: \{\}\\
Caption 2: \{\}\\
\\
You must keep your thinking out loud and respond in less than 200 words. Please output your judgment at the end using the following format:
\\
Better Caption: Caption 1 or Caption 2
\end{tcolorbox}

We provide an example of GPT-4V's response for 3D captioning evaluation, corresponding to Fig.~\ref{fig:supp-3d} (b), where caption 1 is Cap3D and caption 2 is our VisualFactChecker.

\begin{tcolorbox}[colback=mplgreen!3,colframe=mplgreen!75!white,title=\textsc{GPT-4V evaluation response (3D objects)},left=0.2ex,right=0.2ex,top=0.2ex,bottom=0.2ex]
\fontsize{7}{7}\selectfont
To evaluate the captions, let's consider the criteria of correctness and detailedness.
\\\\
Caption 1 is very basic, only stating that the object is a ``3D model of a racing car''. This is correct but lacks detail.
\\\\
Caption 2, on the other hand, gives a much more comprehensive description. It addresses the color scheme (``white and red sports car, with a black and red stripe''), the decorative elements (``checkered pattern on the hood''), specific parts of the car (``spoiler on the back'', ``curved roof'', ``black and sporty tires''), and even the background which helps to highlight the object (``positioned on a neutral background''). This caption goes beyond simply identifying the object as a racing car; it describes its aesthetic and functional elements, which aligns with both correctness and detailedness.
\\\\
While Caption 1 is correct, it is very brief and does not provide any detail. Caption 2 is also correct but provides a rich and detailed description that would give someone who cannot see the image a clear understanding of the 3D model's appearance and character.
\\\\
Better Caption: Caption 2
\end{tcolorbox}

\section{More Qualitative Results}
\label{appendix:vis}
\begin{figure*}[t]
\centering
\includegraphics[width=0.7\textwidth]{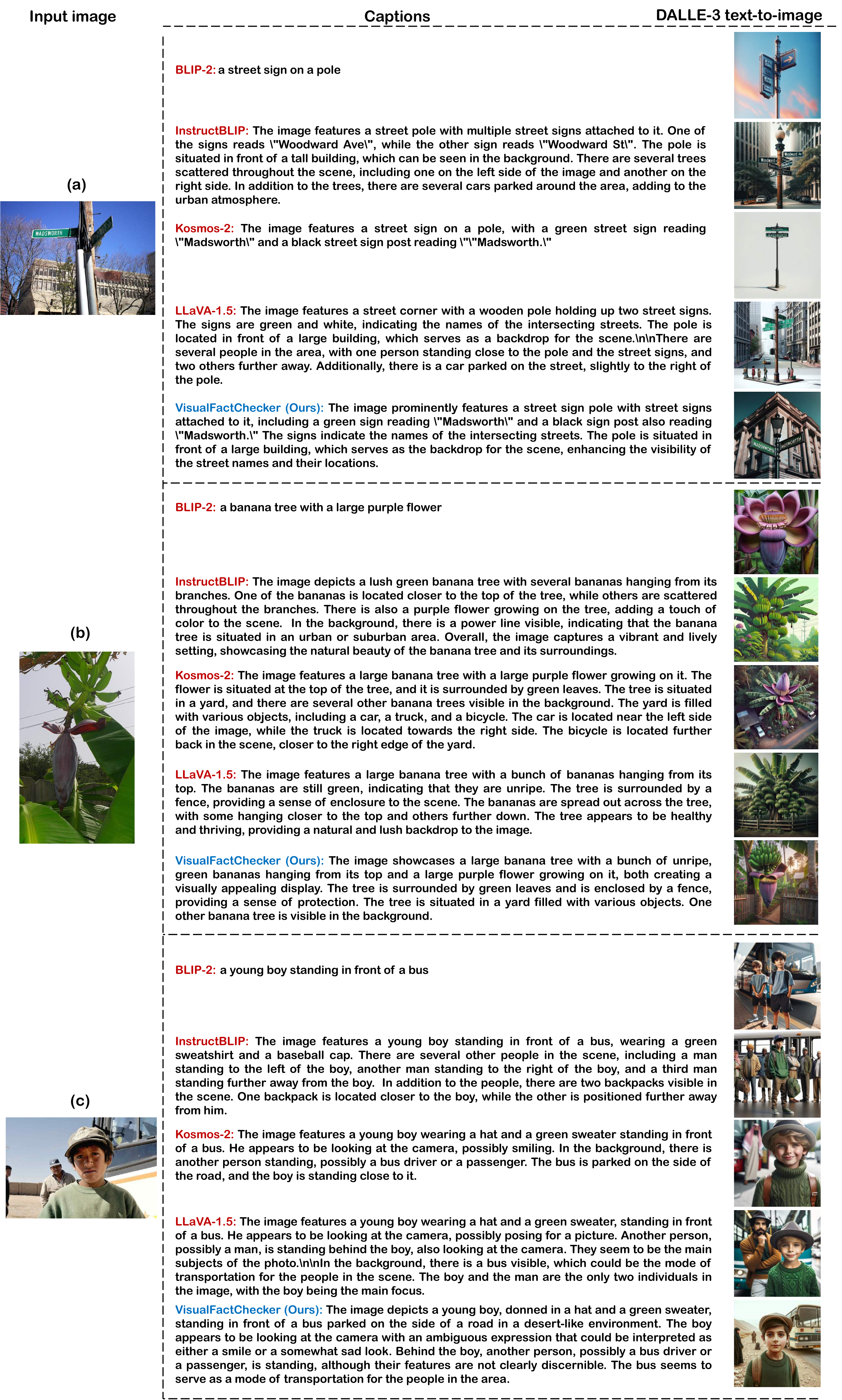}
\caption{Comparison of different 2D image captioning methods (1/3). The left column is the input image from COCO, the middle column shows the caption generated by different methods (BLIP-2, InstructBLIP, Kosmos-2, LLaVA-1.5, and our proposed VisualFactChecker). In the right column, we use DALLE-3 as a text-to-image model to reconstruct 2D images using different generated captions.
}
\label{fig:supp-2d-vis-1}
\end{figure*}

\begin{figure*}[t]
\centering
\includegraphics[width=0.75\textwidth]{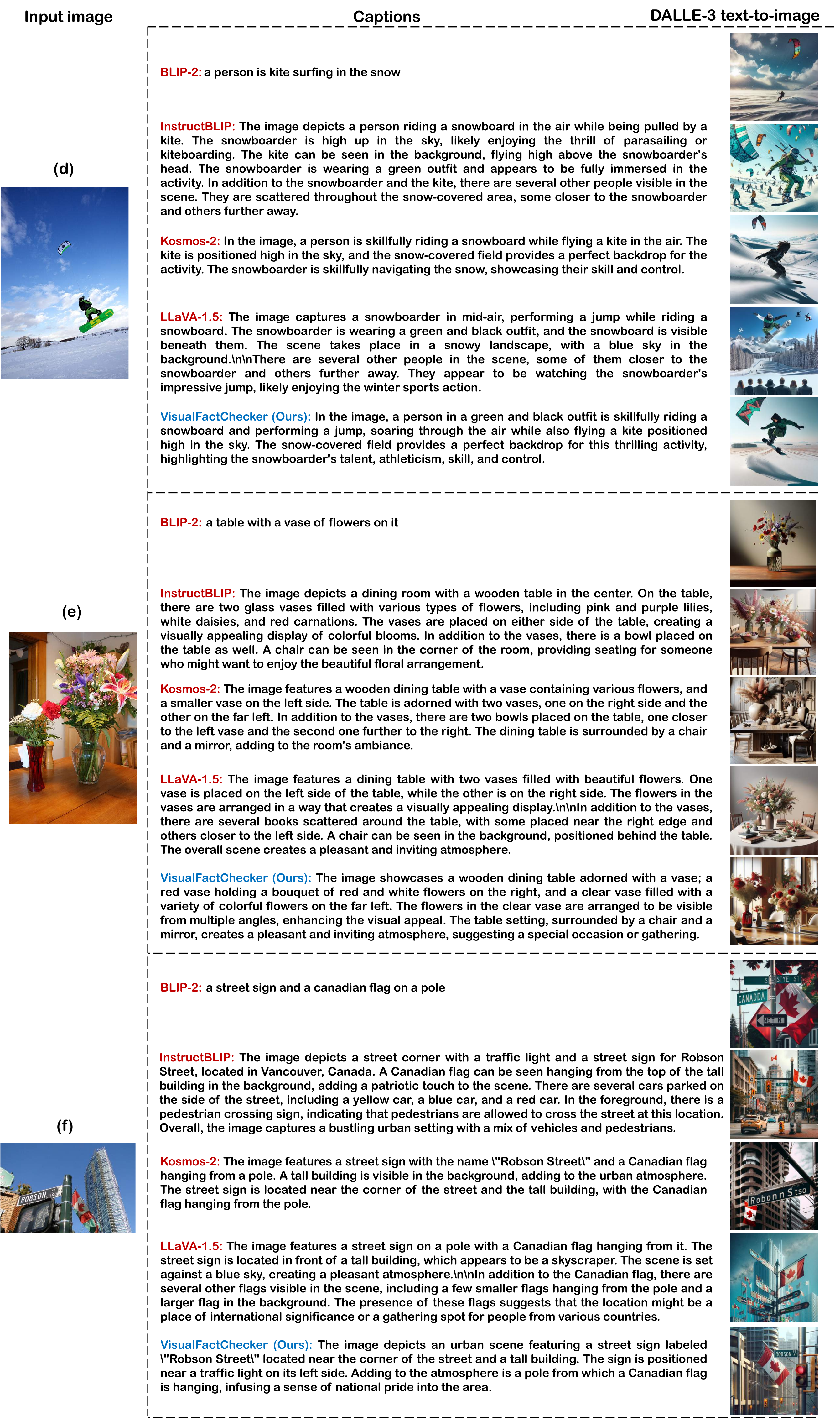}
\caption{Comparison of different 2D image captioning methods -- more examples (2/3). 
}
\label{fig:supp-2d-vis-2}
\end{figure*}

\begin{figure*}[t]
\centering
\includegraphics[width=0.75\textwidth]{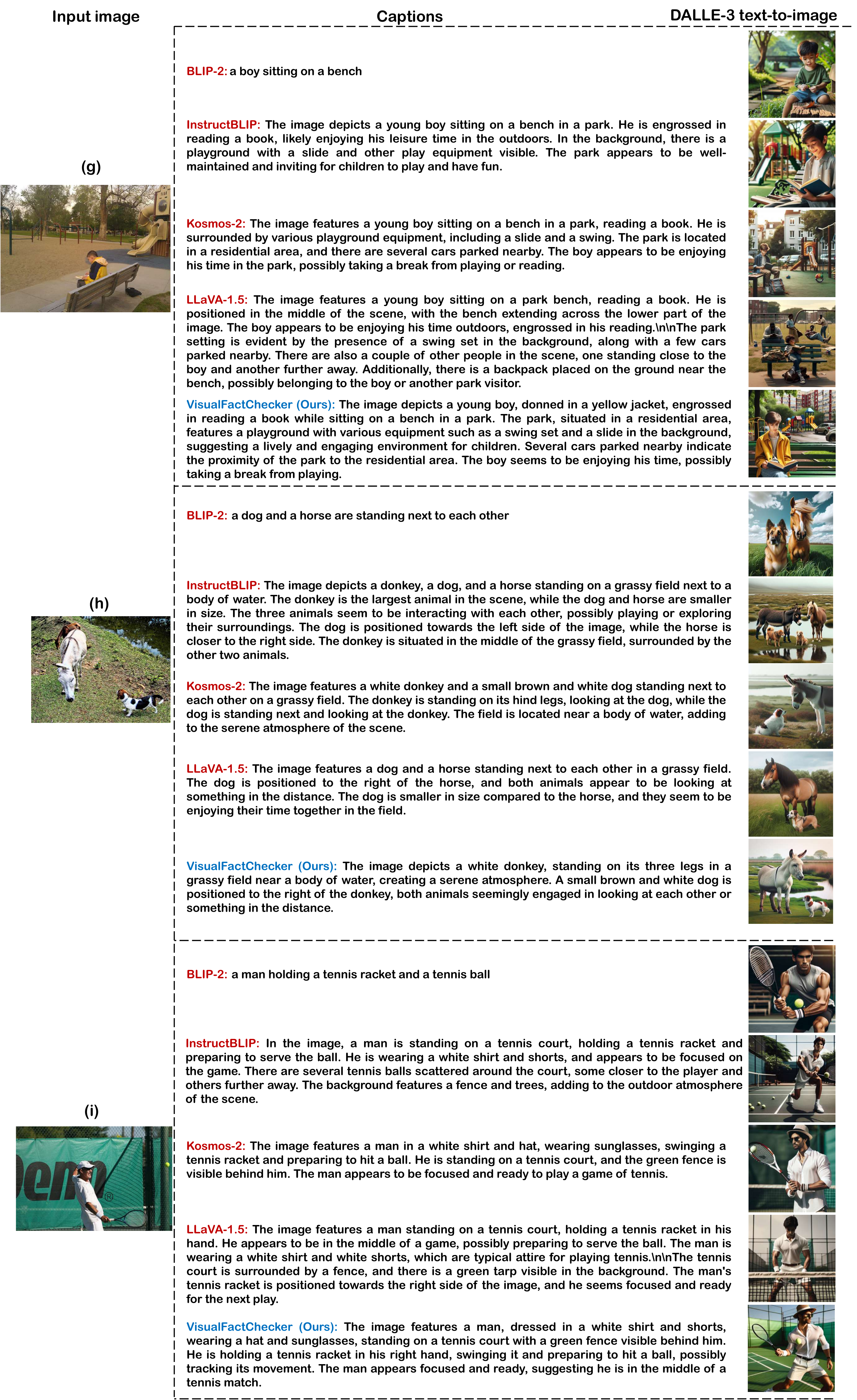}
\caption{Comparison of different 2D image captioning methods -- more examples (3/3). 
}
\label{fig:supp-2d-vis-3}
\end{figure*}

\begin{figure*}[t]
\centering
\includegraphics[width=\textwidth]{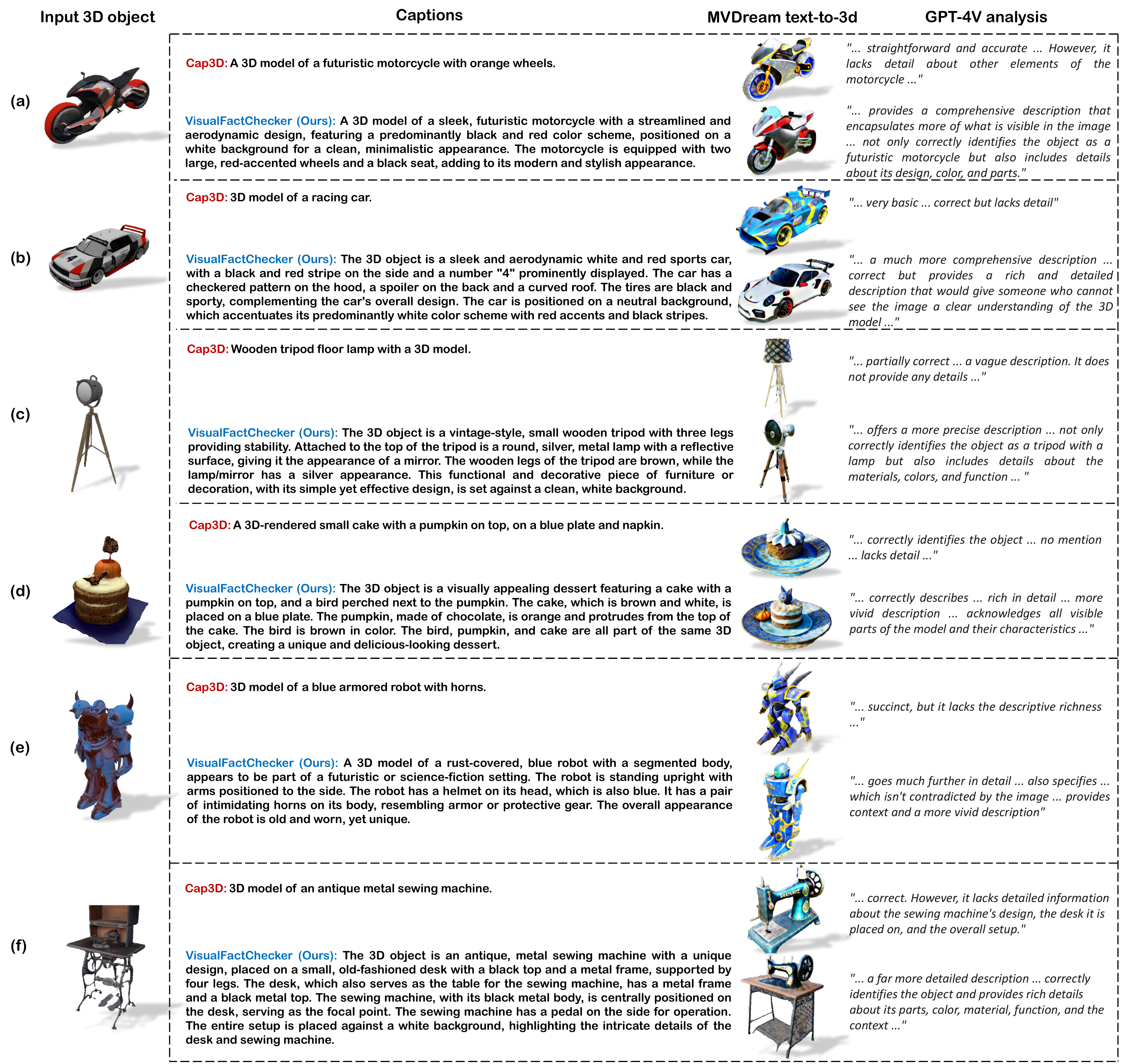}
\caption{Comparison of different 3D object captioning methods. The left column shows the input 3D objects, the middle column shows the captions generated from Cap3D and our VisualFactChecker. In the right column, we use MVDream~\citep{shi2023mvdream} as a text-to-3D model to reconstruct 3D objects using different 3D captions (generated by Cap3D~\citep{luo2023cap3d} and ours). From the results, we can see that the reconstructed 3D objects using Cap3D captions are less similar than the input ones compared with ours, suggesting their captions may not contain sufficient information to reconstruct the 3D object or incorrectly describe the visual contents. 3D objects reconstructed using our captions are more similar to the inputs.}
\label{fig:supp-3d}
\end{figure*}

\begin{figure*}[t]
\centering
\includegraphics[width=\textwidth]{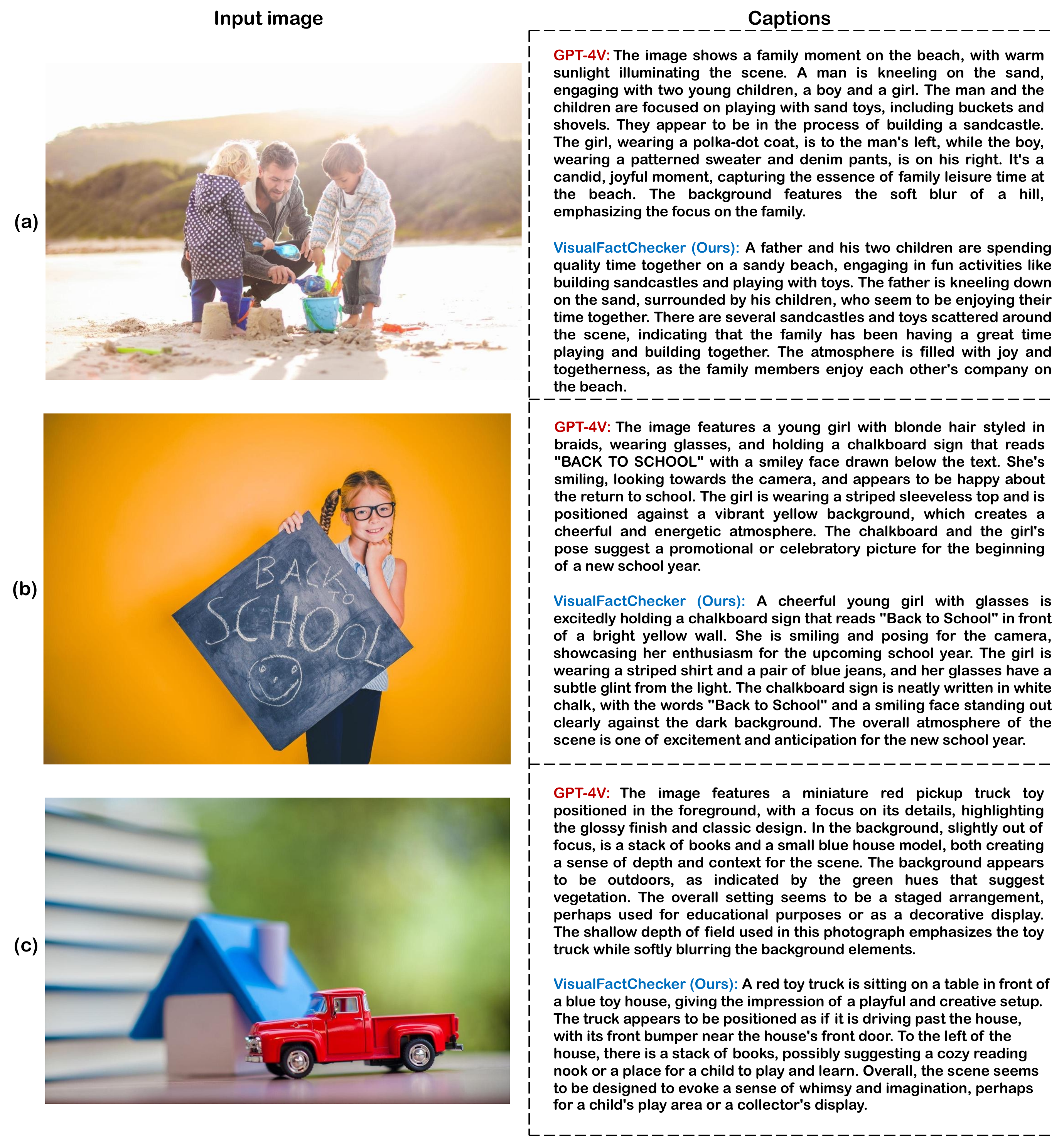}
\caption{Comparison of VisualFactChecker with GPT-4V. Our method can generate high-fidelity detailed captions that closely match GPT-4V’s quality. To compare with GPT-4V, we use Llama-2 as our LLM instead of GPT-4 when generating captions for the above images.
}
\label{fig:supp-getty}
\end{figure*}

We show more results of image captioning methods and their DALLE-3 reconstructed images using different generated captions (COCO 2D images in Fig.~\ref{fig:supp-2d-vis-1},~\ref{fig:supp-2d-vis-2},~\ref{fig:supp-2d-vis-3}; Objaverse 3D objects in Fig.~\ref{fig:supp-3d}).
We show more comparison with GPT-4V captions using Llama-2 as the LLM in Fig.~\ref{fig:supp-getty}.

\section{Following Complex Prompts}
\label{appendix:rewrite}

\begin{figure*}[t]
\centering
\includegraphics[width=\textwidth]{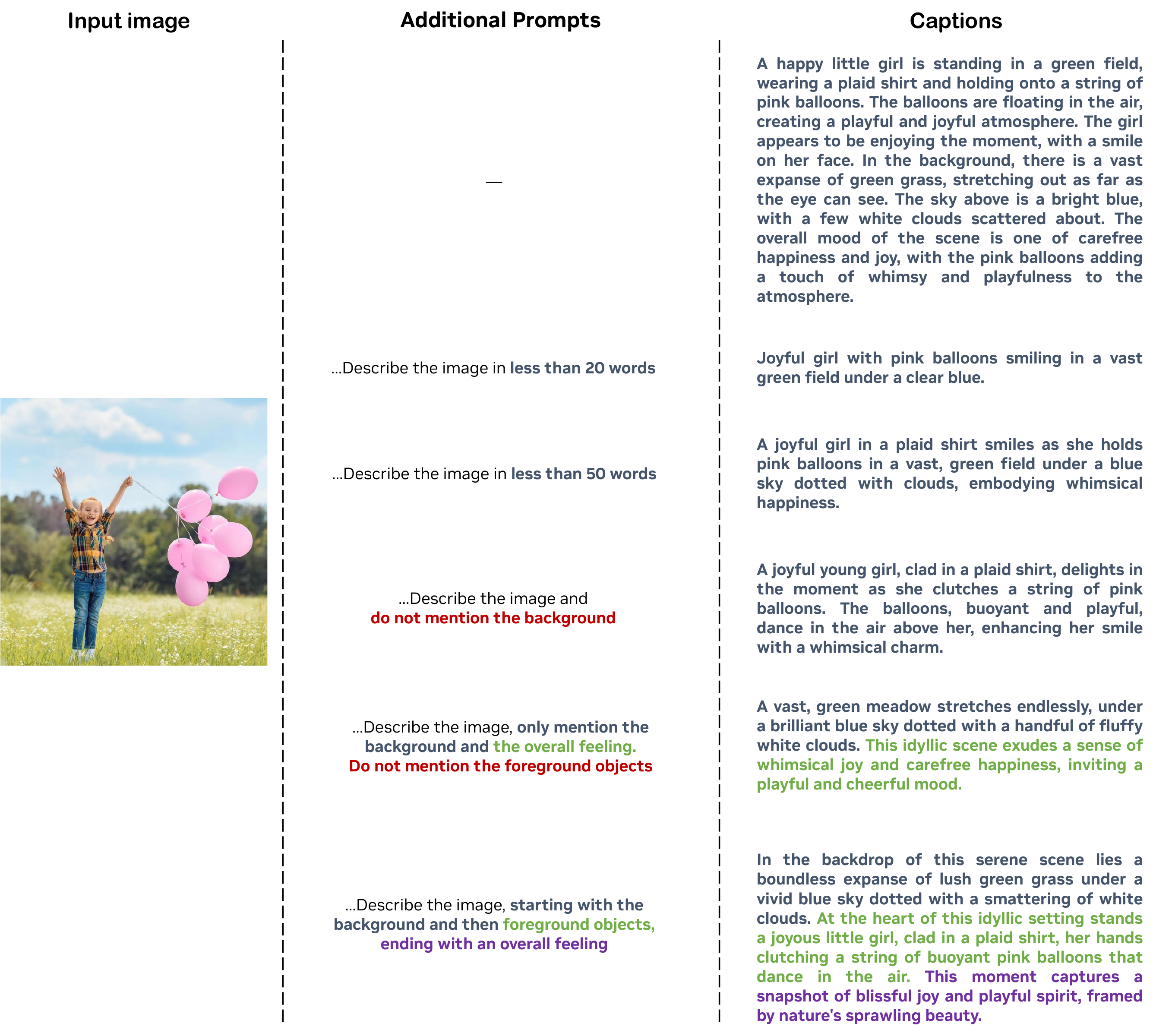}
\caption{Following complex prompts. By leveraging an LLM to write the final caption, VisualFactChecker can follow complex instructions to write captions in various styles.
}
\label{fig:supp-instruction}
\end{figure*}

By leveraging the LLM, VisualFactChecker can follow
complex prompts to write captions in different styles. Examples shown in Fig.~\ref{fig:supp-instruction}.

\end{document}